\documentclass[conference]{IEEEtran}
\IEEEoverridecommandlockouts
\usepackage{cite}
\usepackage{amsmath,amssymb,amsfonts}
\usepackage{algorithmic}
\usepackage{graphicx}
\usepackage{textcomp}
\usepackage{xcolor}
\usepackage{float}
\usepackage{url}

\usepackage{booktabs}

\makeatletter
\let\NAT@parse\undefined
\makeatother
\usepackage{hyperref}  

\usepackage{geometry}
\def\BibTeX{{\rm B\kern-.05em{\sc i\kern-.025em b}\kern-.08em
    T\kern-.1667em\lower.7ex\hbox{E}\kern-.125emX}}
\begin{document}

\newgeometry{left=54pt, right=54pt, top=72pt, bottom=54pt}

\title{Towards Interactive Autonomous Vehicle Testing: Vehicle-Under-Test-Centered Traffic Simulation\\





\thanks{{This work was supported in part by the National Natural Science Foundation of China 52125208, 52232015.

*Corresponding author: Xiaocong Zhao (E-mail: zhaoxc@tongji.edu.cn)
}}
}

\author{\IEEEauthorblockN{Yiru Liu}
\IEEEauthorblockA{
\textit{Key Laboratory of Road and} \\
\textit{Traffic Engineering,} \\
\textit{Ministry of Education,} \\
\textit{Tongji University}\\
Shanghai, China \\
yiru\_liu@tongji.edu.cn}
\and
\IEEEauthorblockN{Xiaocong Zhao*}
\IEEEauthorblockA{
\textit{Key Laboratory of Road and} \\
\textit{Traffic Engineering,} \\
\textit{Ministry of Education,} \\
\textit{Tongji University}\\
Shanghai, China \\
zhaoxc@tongji.edu.cn}
\and
\IEEEauthorblockN{Jian Sun}
\IEEEauthorblockA{
\textit{Key Laboratory of Road and} \\
\textit{Traffic Engineering,} \\
\textit{Ministry of Education,} \\
\textit{Tongji University}\\
Shanghai, China \\
sunjian@tongji.edu.cn}
}

\maketitle

\begin{abstract}
The simulation-based testing is essential for safely implementing autonomous vehicles (AV) on roads, necessitating simulated traffic environments that dynamically interact with the Vehicle Under Test (VUT). This study introduces a VUT-Centered environmental Dynamics Inference (VCDI) model for realistic, interactive, and diverse background traffic simulation. Serving the purpose of AV testing, VCDI employs Transformer-based modules in a conditional trajectory inference framework to simulate VUT-centered driving interaction events. First, the VUT future motion is taken as an augmented model input to bridge the action dependence between VUT and background objects. Second, to enrich the scenario diversity, a Gaussian-distributional cost function module is designed to capture the uncertainty of the VUT's strategy, triggering various scenario evolution. Experimental results validate VCDI's trajectory-level simulation precision which outperforms the state-of-the-art trajectory prediction work. The flexibility of the distributional cost function allows VCDI to provide diverse-yet-realistic scenarios for AV testing. We demonstrate such capability by modifying the anticipation to the VUT's cost-based strategy and thus achieve multiple testing scenarios with explainable background traffic evolution. Codes are available at \url{https://github.com/YNYSNL/VCDI}.

\end{abstract}


\section{Introduction}
The deployment of autonomous vehicles (AV) on public roads is forming a human-machine mixed traffic environment. AVs, as new traffic participants, are found to possibly compromise road safety when interacting with human road users \cite{markkula2020defining, liu2021crash}. 
According to six years (2015-2020) of crash data from the California Department of Motor Vehicles, AVs are involved in collisions four times more frequently than human-driven vehicles\cite{mccarthy2022autonomous}. Despite traffic rules often assigning fault to human drivers, AVs are frequently criticized for their unpredictable interaction behaviors\cite{liu2021crash, mccarthy2022autonomous}. The human-machine mixed driving environment poses new challenges for the interactive capabilities of autonomous driving and also brings new requirements for testing these systems.

AV testing is fundamentally tied to realistic, interactive, and diverse simulation environments where background dynamic participants interact with the vehicle under test (VUT), namely the AV. In pursuing the driving behaviors of background traffic participants that are consistent with real-world traffic, efforts have been widely rolled out on the following challenges, realism of the scene setting, inter-agent interactivity, and diversity of scenario evolution. First, the realism of the scene setting ensures that the testing scenario is possibly encountered in real-world traffic, thus providing testing value. Second, inter-agent interaction calls for congruent action-dependence among VUT and background traffic participants. Third, the diversity of scenario evolution is essential as it reflects the inherent uncertainty and dynamics of real-world traffic, ensuring that the simulation environment authentically represents the wide range of situations VUT might encounter during normal operations.

\begin{figure*}[!ht]
  \centering
  \includegraphics[width=0.9\textwidth]{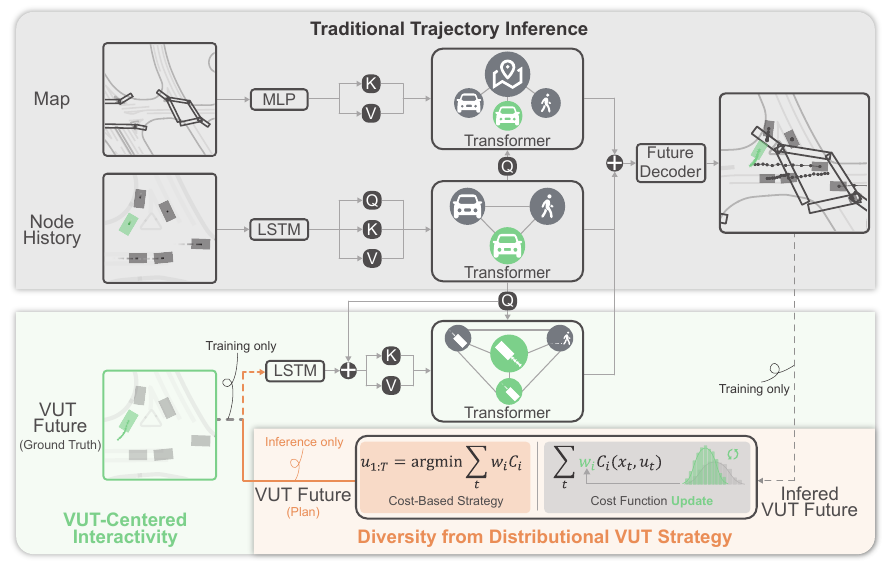}
  \caption{VUT-Centered Environmental Dynamics Inference Framework.}\label{figure:1}
\end{figure*}

Focusing on varied points, solutions have been developed to generate testing environments in the simulation while handling the above challenges. The log test is primarily applied to ensure realism by directly replaying the real-world trajectories performed by human drivers while suffering from the lack of interaction with VUT \cite{nalic2020scenario}. By introducing behavior models like the Intelligent Driver Model (IDM) to enable background participants' responses to VUT \cite{ciuffo2014global}, the model-based test advances the testing environments to be dynamic. While a step towards interactivity, the model-based test often results in overly cautious background object behaviors, avoiding collisions and failing to replicate high-risk scenarios that are crucial for testing. The learning-based test generates interaction-aware background participants by learning implicit interactive relationships from driving data \cite{chen2024data}. However, the interactivity does not extend to VUT which means background participants take VUT as a non-strategic moving obstacle, isolating VUT from the multi-agent interaction.

Serving the purpose of trustworthy AV testing, this work proposes a VUT-Centered environmental Dynamics Inference (VCDI) framework for simulating background traffic participants that interact with VUT in a human-like manner. Specifically, VCDI applies a Transformer-based neural network model to infer the future trajectories of traffic participants (Fig. \ref{figure:1}). We initialize the simulation with real-world driving scenario records to ensure the realism of the scenario setting. To capture background traffic objects' responses to the VUT, the anticipated trajectory of VUT is taken as additional model input, facilitating VUT-centered interactivity. For the application of AV testing where VUT's strategy is unavailable, the strategy presets of VUT are generated by a cost function based on Gaussian distribution, which is learned from real-world AV-involved driving data. By varying the strategy presets, VCDI could simulate diverse-yet-realistic background traffic dynamics from real-world initial scenario states.

The contributions of this work are three-fold:
\begin{itemize}
\item A conditional trajectory inference framework VCDI is proposed to model background traffic objects' strategic responses, centering simulated environmental dynamics on the VUT.
\item A Gaussian-distributional cost function module is designed and integrated to provide flexible and explainable inference conditions, enabling diverse traffic simulation.
\item VCDI's potential for directional background traffic simulation is demonstrated through experiments in typical traffic scenes.
\end{itemize}


The remainder of this work starts from a review of related works (Section \ref{sec:2}) which systematically analyzes the pros and cons of background traffic simulation techniques for AV testing from previous efforts. We then introduce the design details of the VCDI framework in Section \ref{sec:3} and the validation settings in Section \ref{sec:4}. Results are discussed in Section \ref{sec:5} to demonstrate VCDI's capability of simulating diverse traffic environments. Finally, we conclude our work in Section \ref{sec:6}.

\section{Related Works}\label{sec:2}
The pursuit of autonomous driving safety through on-road testing presents substantial challenges and costs. It has been estimated that billions of kilometers are required to validate the safety of AV, an endeavor that is impractical in terms of both time and resources \cite{kalra2016driving}. Consequently, the focus has shifted towards simulation-based methodologies where challenges lie in the reproduction of real-world traffic that interacts properly with VUT \cite{nalic2020scenario}. 

The log test, which replays recorded trajectory data from actual driving scenarios, offers high reproducibility, enabling researchers to consistently test VUT under controlled conditions \cite{arief2018accelerated,li2024scenarionet,gulino2024waymax}. As an extension, Wang et al. extract principle components of the traffic, termed traffic primitives, from the logged data to guide new scenario generation, enriching the log test from being log-only \cite{wang2018extracting}. However, the static nature of the log test is a major drawback. Since all scenarios are predetermined, the log test fails to account for the dynamic nature of real-world driving where traffic conditions are constantly affected by VUT.

To address the non-interactive limitations of the log test, analytical models \cite{saifuzzaman2014incorporating,zheng2014recent} like the Intelligent Driver Model (IDM) were applied \cite{treiber2000congested}, constructing the model-based test \cite{kusari2022enhancing}. These models allow background objects to respond to VUT's actions, providing a certain level of interaction. However, despite the introduction of interactivity, analytical models often result in overly cautious behavior from background vehicles, thus failing to simulate hazardous interaction patterns. The behaviors of background traffic participants in the model-based test are still bounded by the model parameters and typically fail to replicate high-risk scenarios that are crucial for exhaustive testing.

Following the identification of limitations inherent in both the log test and the model-based test, the research community has sought integrated approaches that combine the inherent realism of the log test with the interactivity of the model-based test. The primary challenge lies in accurately replicating complex and multi-objective scenarios that not only demand a high degree of realism but also necessitate intricate interactions and strategic considerations among traffic participants. 

Deep learning (DL) models emerge as a viable methodology in addressing these challenges, which can incorporate implicit interactive relationships among traffic participants \cite{nalic2020scenario,tan2021scenegen,suo2021trafficsim,xu2023bits}. The key advantage of leveraging such implicit interactions lies in their ability to simulate a more realistic traffic environment, where background vehicles can exhibit complex and human-like driving behaviors. Mütsch et al. have noted an emerging trend where DL models are progressively supplanting traditional model-based simulation in the fields of autonomous driving testing \cite{mutsch2023model}. DL-based simulation models, despite their capabilities of capturing complex traffic patterns, primarily focus on the interaction among modeled agents, i.e., the background objects while isolating the VUT. Therefore, the effectiveness of the DL-based test is confined to reacting to observable VUT states, lacking the foresight to interpret and adapt to VUT's underlying strategic objectives. 

The progression from log test to model-based test and to DL-based test reflects the ongoing need for AV testing scenarios that are both realistic and interactive. In this work, we incorporate the strategy presets of VUT as inputs that influence the future trajectory inference of background participants. Concurrently, the strategy presets of VUT act as the keys to initiating scenarios. Different presets for VUT unlock varied scenario evolution, thereby providing a dynamic and adaptable testing environment.

\section{Method}\label{sec:3}
\subsection{Problem Description}\label{AA}
Our work is based on a DL-based trajectory prediction model \cite{huang2023differentiable} where trajectories of background traffic participants are inferred over a forecast horizon $T_{f}$. In addition to the historical states of all agents and environmental context, we further applied the anticipated future trajectory of VUT as input to capture background traffic participants' responses to the strategy of VUT. Let $S_{VUT}^{fut}$ represent the future trajectory of VUT, which is a sequence of anticipated states over the forecast horizon. The input $X$ to the inference model includes, along with the current environmental information $M$, the historical states over a time period $T_{h}$ of VUT $S_{0}$, the historical states of $N$ surrounding agents $S_{1}, S_{2}…S_{N}$, and the future trajectory of VUT $S_{VUT}^{fut}$. This enhancement aims to capture the reactive and strategic adjustments of surrounding agents to VUT's future motion. The output of the trajectory inference model $Y$ is a trajectory set consisting of a sequence of 2D coordinates denoting the possible future position of the background traffic participants. The problem is formally expressed as:
\begin{equation}\label{eq:1} 
\begin{aligned}
X&=\left\{M,S_{0},S_{1},S_{2}…S_{N},S_{VUT}^{fut}\right\} \\
Y&=\left\{\left(x_{i}^{t}, y_{i}^{t}\right) \mid t \in\left\{t_{0}+1, \ldots, t_{0}+T_{f}\right\}\right\}_{i=1}^{N}
\end{aligned}
\end{equation}
where $S_{i}=\left\{s_{i}^{t_{0}-T_{h}+1}, s_{i}^{t_{0}-T_{h}+2}, \ldots, s_{i}^{t_{0}}\right\}$ is the dynamic state of the agent $i$ at timestep $t$, $(x_{i}^{t}, y_{i}^{t})$ is the inferred coordinate of the target agent $i$ at timestep $t$, and $t_{0}$ is the current time step \cite{huang2022recoat}.

\subsection{VUT-Centered Environmental Dynamics Inference}

As shown in Fig. \ref{figure:1}, VCDI incorporates a Transformer-based neural network model for trajectory inference. The model processes the static input of the map through a Multilayer Perceptron (MLP), encoding the geometric and infrastructural context, and the dynamic input comprising the node history via a Long Short-Term Memory (LSTM) network, encapsulating the participants' movement patterns. Additionally, the future trajectory of VUT is encoded using an LSTM network. These processed inputs are fed into the Transformer network, which captures spatial-temporal relationships between the elements within the traffic scene. Detailed specifications of the framework are outlined below.

\subsubsection{Traditional Trajectory Inference}
Traditional trajectory inference primarily incorporates the historical states of traffic participants into feature input and encoding, failing to consider the impact of the VUT's anticipated behaviors on the inference process, as shown below.

\paragraph{Input Representation} Each agent's history is captured as a sequence of dynamic features over the past 20 timesteps ($T_{h}$=20), including 2-D position, heading angle, velocity, and bounding box size. We focus on the nearest 10 agents ($N$=10) to VUT, storing their observations in a fixed-size tensor, padding absent agents with zeros. Scene context is derived from lanes and crosswalks, represented by vectors of waypoints with varying features. For each agent, a local scene context is constructed, highlighting probable lanes and nearby crosswalks. Lane waypoints feature center and boundary positions, headings, speed limits, and traffic signals. Crosswalk waypoints detail the perimeters of the crosswalk areas. Positional attributes of all agents and map elements are recalibrated to the VUT's local coordinate frame. 

\paragraph{Encoding Scene Context and Historical States} The scene context is processed using an MLP, which involves encoding nearby lanes and crosswalks into vectors. These vectors represent numerical features such as positions and directions. The MLP enables discrete feature encoding and is essential for capturing the nuanced spatial relationships that influence an agent’s behavior. In addition, our model adopts an LSTM network to encode the historical states of agents, optimizing for predictive performance and computational efficiency, particularly in scenarios involving short-term time series data. 

\paragraph{Modeling the Interactions between Agents and Scenes} We construct an agent-agent interaction graph and an agent-scene interaction graph to capture the interaction relationships. Specifically, we use a two-layer self-attention Transformer encoder to process the agent-scene interaction graph, where the query, key, and value are the encoded history states. Furthermore, we utilize two cross-attention Transformer encoders for the agent-scene interaction graph, with agents' interaction features serving as the query, and map features as the key and value. These graphs facilitate the understanding of complex dynamics and dependencies within the scene.

\paragraph{Decoding Futures} The decoding phase involves translating the interaction encodings into multimodal future trajectories for each agent. This is achieved by concatenating the agent-agent and agent-map interaction encodings, and processing them through an MLP to predict potential future states. The model is designed to output the trajectory with the highest probability for each agent as the final inferred trajectory.

\subsubsection{Modeling VUT-Centered Interactivity}
The unique setup of our model, as illustrated in Fig. \ref{figure:1}, is the anticipated behaviors of VUT. Specifically, during the training phase, our model utilizes the actual future trajectory sequence of VUT as an additional input. This inclusion enables the model to infer the future trajectories of the surrounding traffic participants as they respond to VUT's strategy. When simulating traffic environments based on the model, instead of using the actual trajectory, we input a planned trajectory by solving a cost-based optimization to handle the inaccessibility of the actual VUT's future trajectory. The planned trajectory is generated based on the cost function learned during training. By sampling different VUT's planned trajectories from the cost function, our model is capable of inferring the corresponding reactive trajectories of the background objects. This allows us to examine how surrounding traffic dynamically adjusts to various strategic presets of VUT, thus providing insights into the interactions between VUT and other road users within the traffic system.
\paragraph{Augmented Input Representation}
Representing the VUT's future trajectory involves a dynamic sequence of features projected across 50 future timesteps ($T_{f}$=50), covering 2-D position, heading angle, velocity, and bounding box size. 
\paragraph{Encoding Anticipated Behaviors of VUT}
To encode anticipated behaviors of VUT, we employ an LSTM network. Our model encodes the actual future trajectory sequences of VUT during training. In contrast, when simulating traffic environments, the planned trajectories obtained via cost-based optimization are encoded instead of the actual trajectories.
\paragraph{Accounting for VUT-centered Interaction} To influence the driving behaviors of surrounding traffic participants with the VUT's anticipated behaviors, we develop a cross-attention Transformer network. We employ the agents' interaction features as the query and combine the VUT's anticipated behavior features with the agents' interaction features, using the combined vector as the key and value. This concatenation enhances the contextual information, potentially allowing the model to capture more complex interaction patterns.

\subsubsection{Distributional Cost Function}

The cost function of the model encompasses various essential driving aspects vital to VUT, such as speed, acceleration, jerk, steering angle, steering change rate, and traffic rule adherence, focusing on staying near the lane centerline and following the lane direction. To promote efficiency, the function incorporates incentives that encourage VUT to maintain speeds within legal limits. Comfort is quantified through metrics evaluating acceleration, jerk, and steering. Moreover, compliance with traffic regulations is secured by imposing penalties for deviations from ideal lane positions and vehicular orientation. The parameters of the cost function are fine-tuned to minimize the divergence between actual trajectories from real-world data and the planning trajectories. The specific design references the methodology outlined in \cite{huang2023differentiable}.

We conceptualize the cost function for VUT as a distribution rather than being single-valued, reflecting the inherent uncertainty of traffic environments. To model and learn the distributional weight parameters of the cost function effectively, we assume that the cost function parameters follow a Gaussian distribution. This approach allows us to capture the essential characteristics of driving behaviors with mean and variance as the primary descriptors. Through forward propagation in the neural network, we learn and optimize the Gaussian parameters, specifically the mean and variance, to fit the data. This process provides us with a distribution of weights for the cost function, which effectively models the variability in driving behaviors.

\subsection{Learning Process}
In constructing the training process for our model, we adopted the loss function framework as described in the referenced work \cite{huang2023differentiable}, adhering to an end-to-end methodology utilizing real-world driving data. The loss function is a composite of four terms, integrating prediction loss for all agents, score loss to calculate the probabilities of different futures, imitation loss, and the cost of planning for VUT. The comprehensive loss equation is defined as:

\begin{equation}\label{eq:2} 
\begin{aligned}
\mathcal{L} = \lambda_1\mathcal{L}_{\text{prediction}} + \lambda_2\mathcal{L}_{\text{score}} + \lambda_3\mathcal{L}_{\text{imitation}} + \lambda_4\mathcal{L}_{\text{cost}}
\end{aligned}
\end{equation}

where $\lambda_1$ through $\lambda_4$ represent scaling weights assigned to these discrete loss components.

\section{Experiments}\label{sec:4}
\subsection{Dataset}
Our research employs the Waymo Open Motion Dataset \cite{ettinger2021large} for training and validating our framework. This extensive dataset is instrumental in providing real-world insights, as it provides plenty of urban driving scenarios characterized by diverse road structures and varied traffic interactions. To dissect a 20-second driving scene from raw data into discrete frames, we establish a 7-second time window, inclusive of a 2-second historical observation span and a subsequent 5-second forecast horizon, sampled at 10 Hz. This window advances through the scene in increments of 10 timesteps (equivalent to 1 second), thereby generating 14 individual segments of 7-second frames.

For the model development, 80\% of these selected scenes are designated for training, constituting a total of 73,304 scenarios. Another 10\%, amounting to 9,163 scenarios, are used for validation, while the remaining 10\% are reserved for testing.

\subsection{Evaluation}
We evaluate the performance of our model with two key metrics: the average displacement error (ADE) and the final displacement error (FDE) over three timesteps (1, 3, and 5 seconds). The metrics are computed for the prediction error of background objects to assess the model's trajectory-level precision.

\subsection{Ablation Study}
An ablation study is conducted to illustrate the specific model components that influence performance.
Compared with the traditional DL-based trajectory prediction methods \cite{cui2019multimodal,espinoza2022deep,huang2023differentiable}, we consider VUT's strategy as an augmented input, which can result in background traffic participants adequately responding to VUT's strategy. To reveal the importance of VUT-centered augmented input, we set up the work in \cite{huang2023differentiable} as a baseline model that drops out the augmented input. As shown in Table \ref{tab:1}, overall, the VCDI outperforms its baseline in the trajectory prediction tasks. 

\begin{table}[H]
    \centering
    \small
    \caption{Ablation Study in Trajectory Inference Task}
    \vspace{-2mm}
    \label{tab:1}
    \begin{tabular}{ccccc}
         \toprule
         \textbf{} & ADE& FDE (1s)& FDE (3s)& FDE (5s)\\
         \midrule
         No Augment & 0.7641 & 0.2423 & 0.8778 & 1.8266 \\
         \textbf{VCDI {\footnotesize s.v.}} & 0.7530 & 0.2113 & 0.8420 & 1.8095 \\
         \textbf{VCDI} & \textbf{0.7461} & \textbf{0.2093} & \textbf{0.8381} & \textbf{1.7767}  \\
        \bottomrule
    \end{tabular}
\end{table}

To capture the diversity and stochastic nature of driving behaviors, we introduced a Gaussian distribution into our model, assuming that the behavior of the VUT follows a distribution rather than a single value. The single-valued version of our model (VCDI s.v.) unifies the behavior of multiple drivers into a fixed parameter, which is highly limiting. In contrast, the model with the distributional feature better reflects the diversity and uncertainty of driving behaviors, thereby significantly enhancing model performance. As shown in Table \ref{tab:1}, the model with the distributional feature performs best, effectively representing the varied nature of real-world driving behaviors. Based on this optimal model performance, we conducted case studies using the learned distribution to further demonstrate the benefits and effectiveness of the VCDI model.


\begin{figure}[!ht]
  \centering
  \includegraphics[width=0.48\textwidth]{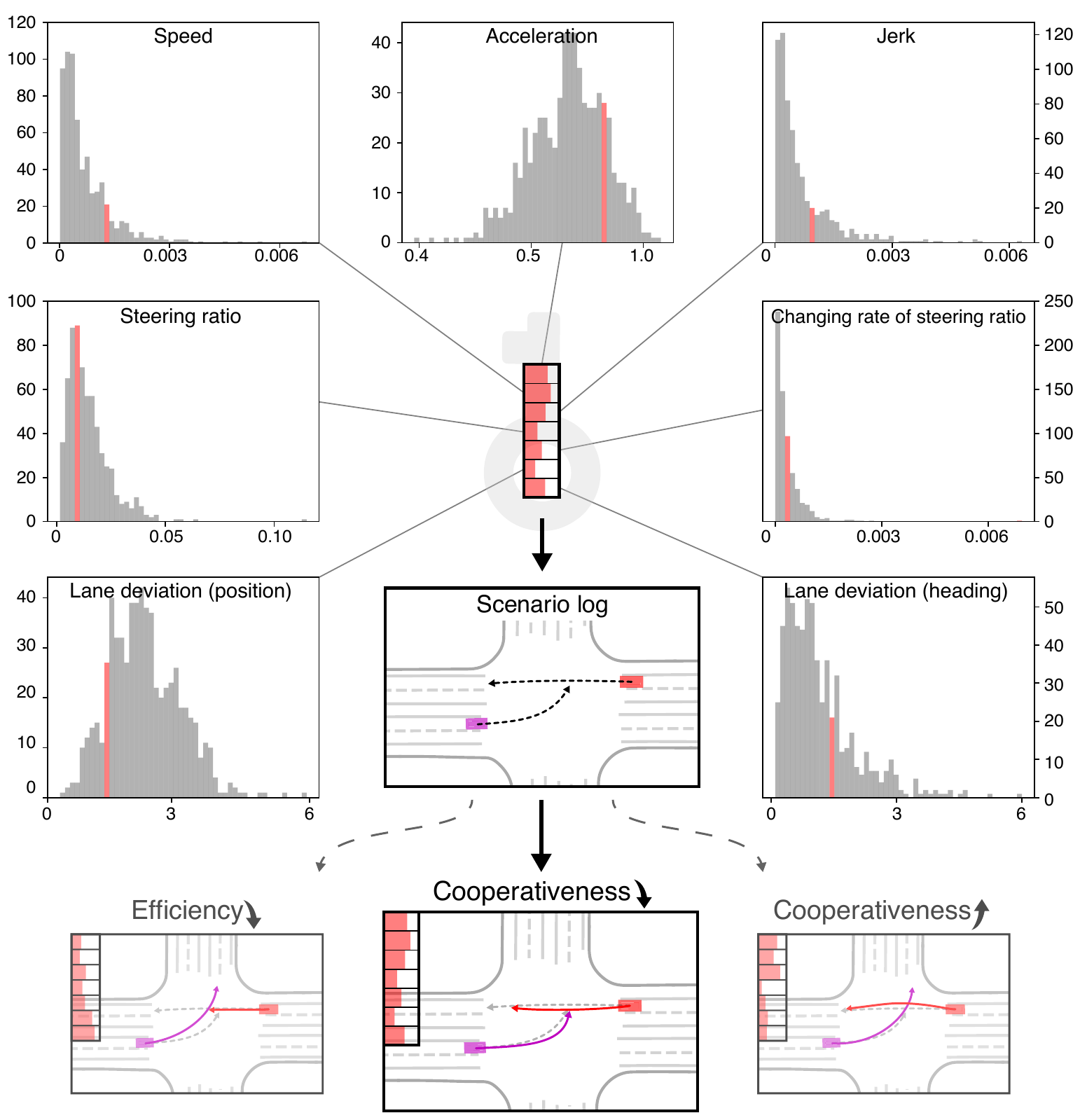}
  \caption{Diverse Scenario Evolution Guided by Distributional Cost Function}\label{figure:2}
\end{figure}

\begin{figure*}[!ht]
  \centering
  \includegraphics[width=0.9\textwidth]{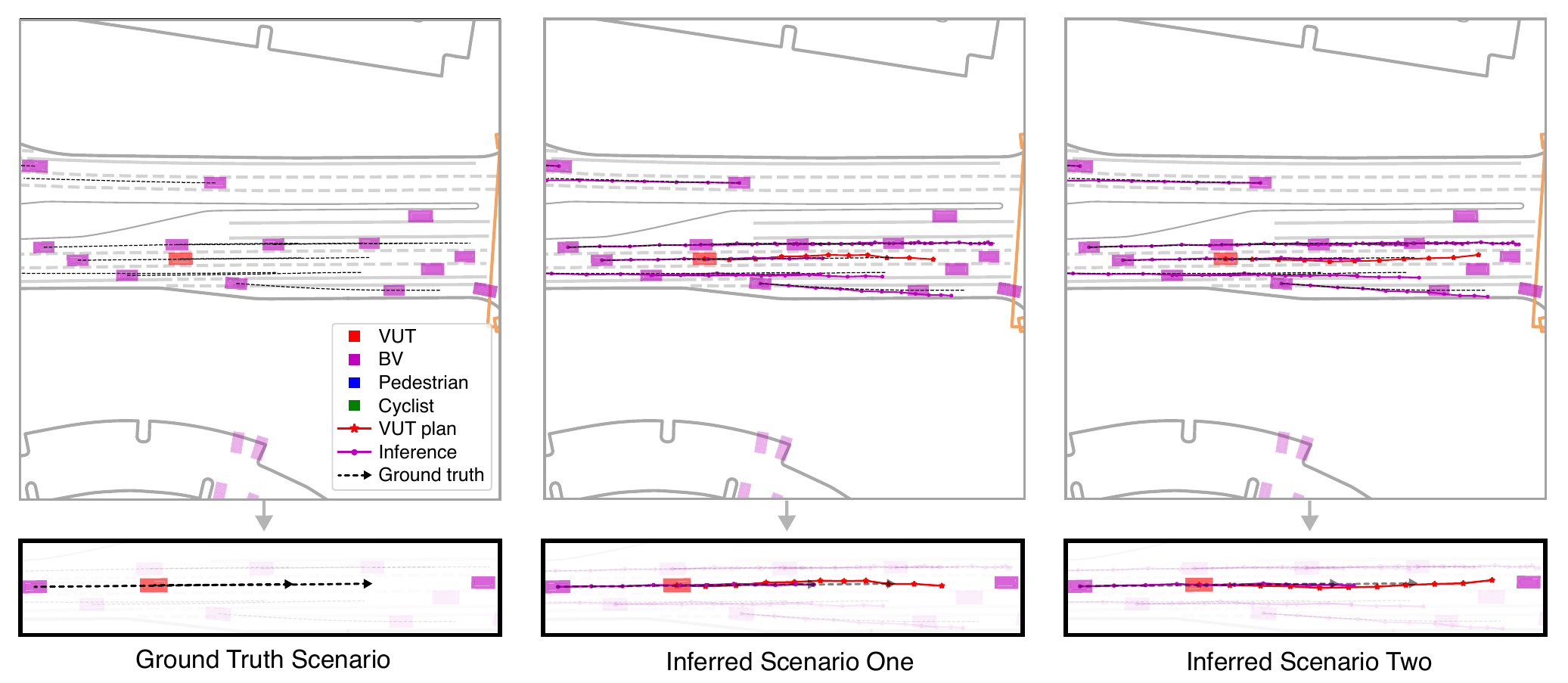}
  \caption{Car-following Scenario}\label{figure:3}
\end{figure*}

\begin{figure*}[!ht]
  \centering
  \includegraphics[width=0.9\textwidth]{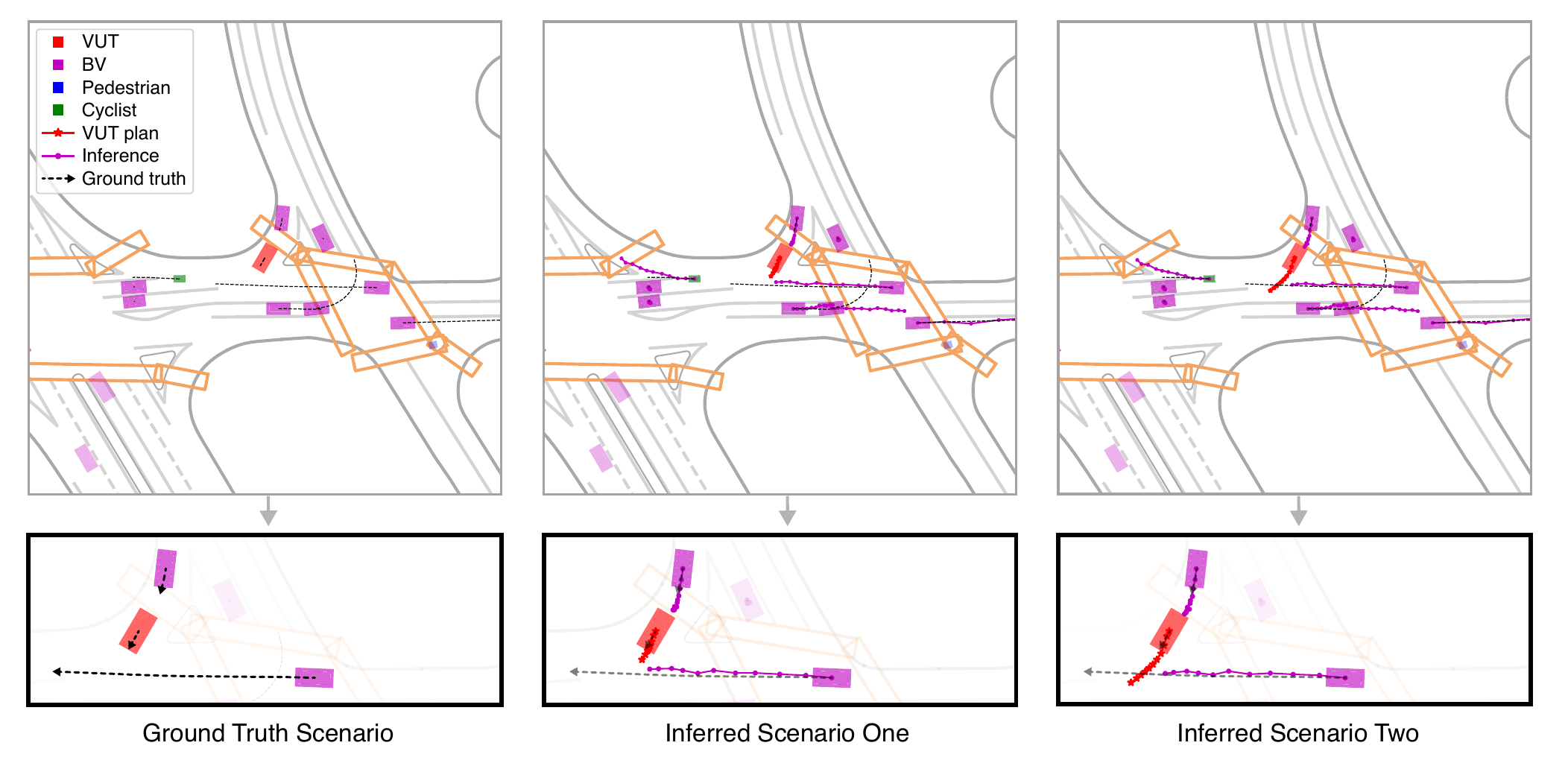}
  \caption{Intersection Scenario}\label{figure:4}
\end{figure*}

\section{Results and Discussion}\label{sec:5}
\subsection{Cost Function Based on Gaussian Distribution}\label{sec:5a}
Fig. \ref{figure:2} illustrates the weights of the distributional cost function generated by our trajectory inference framework. A specific cost function, described by a set of weights from the distribution, serves as a crucial "key" for unlocking scenarios. We could infer scenario evolution with keys either by sampling from the distribution or by making targeted selections.

Specifically, each key, whether sampled or selectively chosen, represents a strategic preset on VUT made by the background objects. The inference results then depict the future trajectories of background objects based on such presets, which allows for simulation that explicitly considers VUT's strategy. In other words, given an initial scenario state extracted from real-world data, the inferred trajectories of background participants change according to the keys used for scenario activation.

\subsection{Typical Scenario Cases}
The distributional cost function obtained from Section \ref{sec:5a} serves as the key to initiating our testing scenarios. From this distribution, we select two different parameter combinations to construct the cost functions, which represent distinct presets of the background traffic participants towards VUT. Subsequently, simulation experiments are conducted in typical traffic scenes extracted from real-world driving data, including the car-following and intersection scenarios.

\subsubsection{Car-following Scenario}

Fig. \ref{figure:3} illustrates both observed and inferred trajectory data. The leftmost subfigure shows actual data from a car-following scenario, while the two subfigures on the right present the trajectories of background participants, inferred based on various strategic presets applied to VUT. Specifically, we increased the emphasis on efficiency in VUT's cost function to varying degrees. As a result of this intensified focus on efficiency, VUT is anticipated to pursue more travel progress, which in turn allows the following car to develop more efficiency-oriented trajectories.

\subsubsection{Intersection Scenario}

Fig. \ref{figure:4} shows data from an intersection scene, where we also incrementally increased VUT’s preference for efficiency. It is evident that when VUT heightened its emphasis on efficiency, there was a noticeable shift in the semantic outcomes. Specifically, as the VUT's focus on efficiency intensified, the behavior of the background vehicle changed from rushing to yielding to the VUT. 

The results of the case study indicate that the different scenario activation triggers varied responses from the background participants. Furthermore, the evolution of simulation scenarios is to some extent controllable, as the key to scenario activation—the distributional cost function—is fully explainable.

\section{Conclusion}\label{sec:6}
The safety of autonomous vehicles on the road requires rigorous testing in environments that closely mimic real-world traffic conditions. This study introduces a VUT-Centered environmental Dynamics Inference (VCDI) framework to simulate background traffic objects. By leveraging a distributional cost function learned from real-world driving data, the VCDI framework uniquely models diverse, strategic responses from surrounding traffic participants to VUT's maneuvers.

Our findings reveal that the VUT-centered augmented input enables background traffic participants, driven by DL-based models, to dynamically interact with VUT. In addition, the cost-based strategy presets on VUT facilitate directional background traffic simulation, meriting the application of AV testing.

\bibliographystyle{IEEEtran}
\bibliography{references}

\end{document}